\documentclass[runningheads]{llncs}

\usepackage{makeidx}  
\usepackage{url}
\usepackage{color}
\usepackage{longtable}
\usepackage{multirow}
\usepackage{graphicx}
\usepackage{subfigure}
\usepackage{amssymb, amsmath, bm}
\usepackage{mathtools}
\usepackage{mathrsfs}
\usepackage{booktabs}
\usepackage{multirow}
\usepackage{cite}
\usepackage{url}
\usepackage{color}
\usepackage{dsfont}
\usepackage[linesnumbered,ruled,vlined]{algorithm2e}
\usepackage[colorlinks,linkcolor=blue]{hyperref}
\usepackage{graphicx}
\usepackage{amsmath,amssymb}
\usepackage{mathrsfs}
\usepackage{makecell}

\usepackage[misc]{ifsym} 
\usepackage{bbding}
\begin{document}

\title{Searching Collaborative Agents for Multi-plane Localization in 3D Ultrasound}
\titlerunning{Searching Collaborative Agents for Multi-plane Localization in 3D US}

\author{Yuhao Huang\inst{1,2}\thanks{Yuhao Huang and Xin Yang contribute equally to this work.}, Xin Yang\inst{1,2\star} \and Rui Li\inst{1,2} \and Jikuan Qian\inst{1,2} \and Xiaoqiong Huang\inst{1,2} \and \\ Wenlong Shi\inst{1,2} \and Haoran Dou\inst{1,2} \and Chaoyu Chen\inst{1,2} \and Yuanji Zhang\inst{3} \and \\ Huanjia Luo\inst{3} \and Alejandro Frangi\inst{1,4,5} \and Yi Xiong\inst{3} \and Dong Ni\inst{1,2}\textsuperscript{(\Letter)}} 


\authorrunning{Huang et al.}

\institute{
\textsuperscript{$1$}School of Biomedical Engineering, Shenzhen University, Shenzhen, China\\
\textsuperscript{$2$}Medical UltraSound Image Computing (MUSIC) Lab, Shenzhen University, China\\
\email{nidong@szu.edu.cn} \\
\textsuperscript{$3$}Department of Ultrasound, Luohu People’s Hosptial, Shenzhen, China\\ 
\textsuperscript{$4$}CISTIB Centre for Computational Imaging \& Simulation Technologies in Biomedicine, School of Computing, University of Leeds, Leeds, UK\\ 
\textsuperscript{$5$}Medical Imaging Research Center (MIRC) – University Hospital Gasthuisberg, Electrical Engineering Department, KU Leuven, Leuven, Belgium \\ }

\maketitle              
\begin{abstract}
3D ultrasound (US) is widely used due to its rich diagnostic information, portability and low cost. Automated standard plane (SP) localization in US volume not only improves efficiency and reduces user-dependence, but also boosts 3D US interpretation. In this study, we propose a novel Multi-Agent Reinforcement Learning (MARL) framework to localize multiple uterine SPs in 3D US simultaneously. Our contribution is two-fold. First, we equip the MARL with a one-shot neural architecture search (NAS) module to obtain the optimal agent for each plane. Specifically, Gradient-based search using Differentiable Architecture Sampler (GDAS) is employed to accelerate and stabilize the training process. Second, we propose a novel collaborative strategy to strengthen agents' communication. Our strategy uses recurrent neural network (RNN) to learn the spatial relationship among SPs effectively. Extensively validated on a large dataset, our approach achieves the accuracy of 7.05$^{\circ}$/2.21\emph{mm}, 8.62$^{\circ}$/2.36\emph{mm} and 5.93$^{\circ}$/0.89\emph{mm} for the mid-sagittal, transverse and coronal plane localization, respectively. The proposed MARL framework can significantly increase the plane localization accuracy and reduce the computational cost and model size.

\end{abstract}

\section{Introduction}
\label{sec:intro}
\begin{figure}[t]
	\centering
	\includegraphics[width=1.0\linewidth]{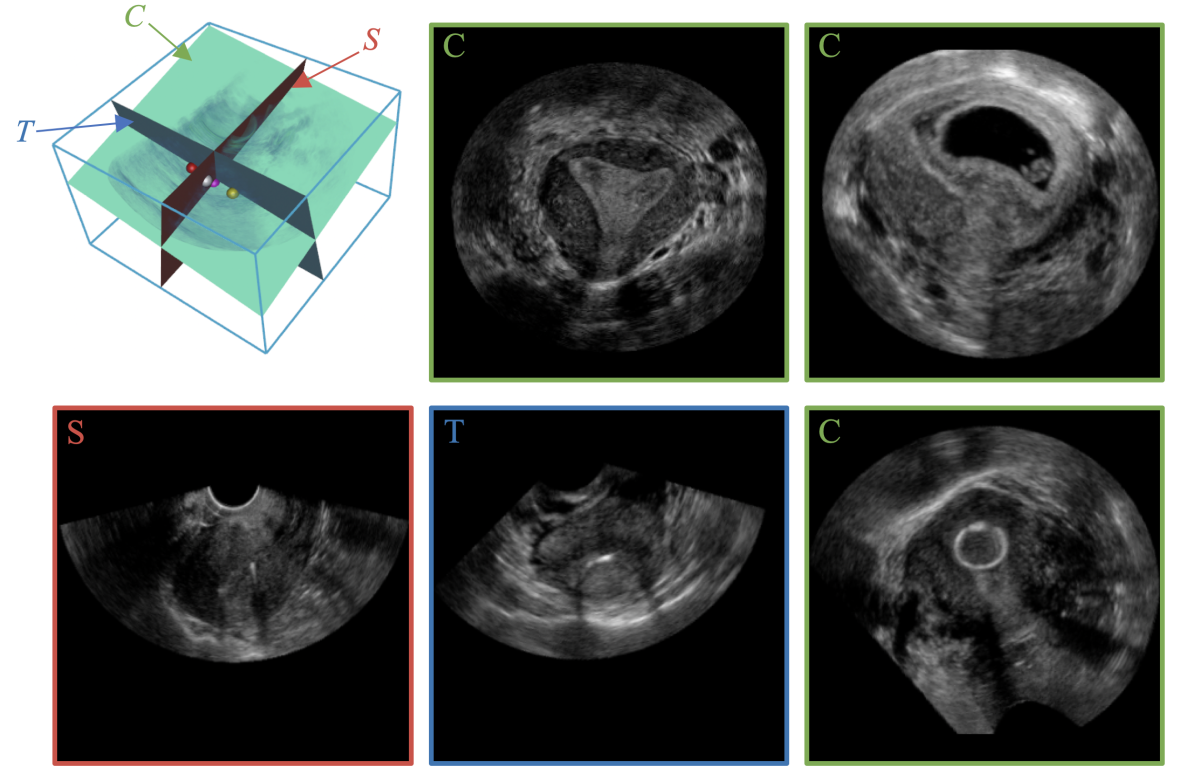}
	\caption{Uterine SPs in 3D US (left to right). \emph{Top row}: spatial layout of SPs, coronal SP of normal and pregnant cases. \emph{Bottom row}: mid-sagittal (S), transverse (T) and coronal (C) SPs of one case with IUD. Red, yellow, purple and white dots are two endometrial uterine horns, endometrial uterine bottom and uterine wall bottom.}
	\label{fig:intro}
\end{figure}

Acquisition of Standard Planes (SPs) is crucial for objective and standardised ultrasound (US) diagnosis~\cite{ni2014standard}. 3D US is increasingly used in clinical practice mainly because of its rich diagnostic information not contained in 2D US. Fig.~\ref{fig:intro} shows coronal SPs that can only be reconstructed from 3D US. They are important for assessing congenital uterine anomalies and Intra-Uterine Device (IUD) localization~\cite{wong2015three}. 3D US can also contain multiple SPs in one shot, which improves scanning efficiency while reducing user-dependence. However, SPs are often located in 3D US manually by clinicians, which is cumbersome and time-consuming owing to a large search space and anatomical variability. Hence, automated SPs localization in 3D US is highly desirable to assist 
in scanning and diagnosis.

As shown in Fig.~\ref{fig:intro}, automatic acquisition of uterine SPs from 3D US remains very challenging. First, the uterine SP often has extremely high intraclass variation due to the existence of IUD, pregnancy and anomaly. Second, the three SPs have very different appearance patterns, which makes designing machine learning algorithms difficult. The third challenge lies in the varied spatial relationship among planes. For 3D uterine US, in most cases, the three planes are perpendicular to each other. However, sometimes, due to uterine fibroids, congenital anomalies, etc., their spatial relationship may be different than expected.

In the literature, plane localization approaches have adopted machine learning methods to various 3D US applications. Random Forests based regression methods were first employed to localize cardiac planes~\cite{chykeyuk2013class}. Landmark alignment~\cite{lorenz2018automated}, classification~\cite{ryou2016automated} and regression~\cite{li2018standard,schmidt2019offset} methods were then developed to localize fetal planes automatically. Recently, Alansary et al.~\cite{alansary2018automatic} first proposed using a Reinforcement Learning (RL) agent for view planning in MRI volumes. Dou et al.~\cite{dou2019agent} then proposed a novel RL framework with a landmark-aware alignment module for effective initialization to localize SPs in noisy US volumes. The experiments have shown that such an approach can achieve state-of-the-art results. However, these RL-based methods still have disadvantages. First, they were only designed to learn a single agent for each plane separately. The agents cannot communicate with each other to learn the inherent and invaluable spatial relationship among planes. Second, agents often use the same network structure such as the VGG model for different planes, which may lead to sub-optimal performance because SPs often have very different appearance patterns.

In this study, we propose a novel Multi-Agent RL (MARL) framework to localize multiple uterine SPs in 3D US simultaneously. We believe we are the first to employ a MARL framework for this problem. Our contribution is two-fold. First, we adopt one-shot neural architecture search (NAS)~\cite{ren2020comprehensive} to obtain the optimal agent for each plane. Specifically, Gradient-based search using Differentiable Architecture Sampler (GDAS)~\cite{dong2019searching} is employed to make the training more stable and faster than Differentiable ARchiTecture Search (DARTS)~\cite{liu2018darts}. Second, we propose a Recurrent Neural Network (RNN) based agent collaborative strategy to learn the spatial relationship among SPs effectively. This is a general method for establishing communication among agents.

\section{Method}
\begin{figure}[htb]
	\centering
	\includegraphics[width=1.0\linewidth]{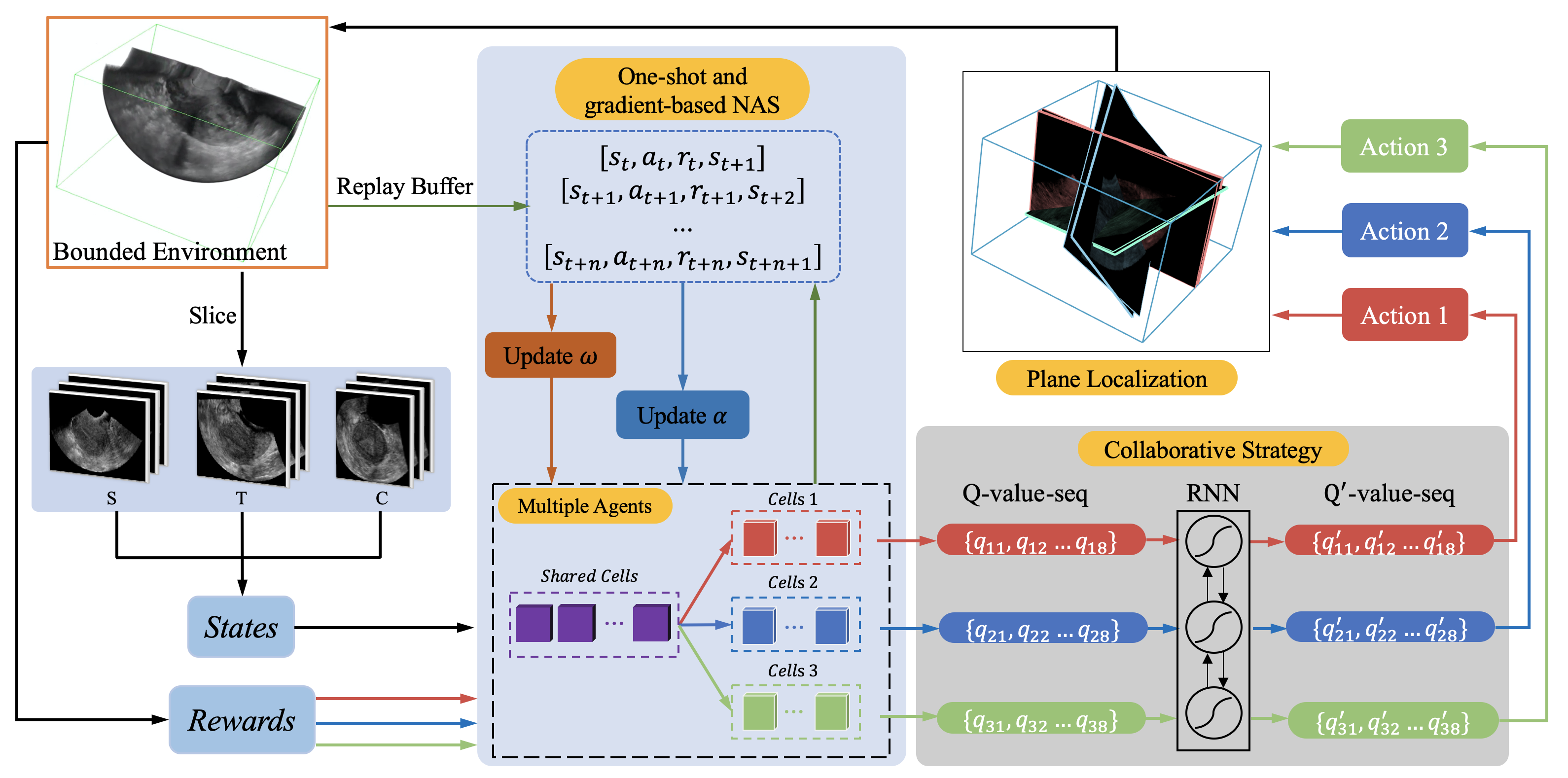}
	\caption{Overview of the proposed framework.}
	\label{fig:framework}
\end{figure}

Fig.~\ref{fig:framework} is the schematic view of our proposed method. We propose a MARL framework to localize multiple uterine SPs in 3D US simultaneously. To improve the system robustness against the noisy US environment, we first use a landmark-aware alignment model to provide a warm start for the agent~\cite{dou2019agent}. Four landmarks used for the alignment are shown in Fig.~\ref{fig:intro}. We further adopt the one-shot and gradient-based NAS to search the optimal agent for each plane based on the GDAS strategy~\cite{dong2019searching}. Then, the RNN based agent collaborative strategy is employed to learn the spatial relationship among planes effectively.

\subsection{MARL Framework for Plane Localization}
To localize multiple uterine planes effectively and simultaneously, we propose a collaborative MARL framework. The framework can be defined by the \emph{Environment}, \emph{States}, \emph{Actions}, \emph{Reward Function} and \emph{Terminal States}. In this study, we take a uterine US volume as the $Environment$ and define $States$ as the last nine planes predicted by three agents, with each agent obtaining three planes. A plane in Cartesian coordinate system is defined as $\cos(\zeta)x+\cos(\beta)y+\cos(\phi)z+d=0$, where $(\cos(\zeta),\cos(\beta),\cos(\phi))$ represents the normal vector and $d$ is the distance from the plane to the volume center. The plane parameters are updated according to its agent's $Actions$ defined as \{$\pm$\emph{\textbf{a}}$_{\zeta}$, $\pm$\emph{\textbf{a}}$_{\beta}$, $\pm$\emph{\textbf{a}}$_{\phi}$, $\pm$\emph{\textbf{a}}$_{d}$\}. Each action taken by each agent will get its own reward signal $R \in \{-1, 0, +1\}$ calculated by the $Reward$ $Function$ $\emph{R}=\emph{sgn}($\emph{D}$($\emph{P}$_{t-1}$$-$\emph{P}$_{g}$$)-$\emph{D}$($\emph{P}$_{t}$$-$\emph{P}$_{g}$$))$, where \emph{D} calculates the Euclidean distance from the predicted plane \emph{P}$_{t}$ to the ground truth plane \emph{P}$_{g}$. For the $Terminal$ $States$, we choose a fixed 50 and 30 steps during training and testing respectively based on our experiments and observation.

The agents aim to maximize both the current and future rewards and optimize their own policies for localizing corresponding SPs. To mitigate the upward bias caused by deep Q-network (DQN)~\cite{mnih2015human} and stabilize the learning process, we use double DQN (DDQN)~\cite{van2016deep} and its loss function is defined as:
\noindent
\begin{equation}
L = E[(r_{t}+\gamma Q(s_{t+1},\mathop {argmax} \limits_{a_{t+1}}Q(s_{t+1},a_{t+1};\omega);\tilde{\omega})-Q(s_{t},a_{t};\omega))^2]
\label{equ0}
\end{equation}
where $\gamma$ is a discount factor to weight future rewards. The data sequence, including states $s_{t}$, actions $a_{t}$, rewards $r_{t}$ at steps $t$ and the next states $ s_{t+1}$, is sampled from the experience replay buffer, which ensures the input data is independent and identically distributed during training. $\omega$ and $\tilde{\omega}$ are the weights of current and target networks. $a_{t+1}$ is the actions in the next step predicted by the current network. Instead of outputting Q-values by fully connected layers of each agent directly and taking actions immediately, here, we propose to use an RNN based agent collaborative module to learn the spatial relationship among SPs.

\subsection{GDAS based Multi-agent Searching}
In deep RL, the neural network architecture of the agent is crucial for good learning performance~\cite{li2017deep}. Previous studies~\cite{alansary2018automatic,dou2019agent} used the same VGG model for localizing different planes, which may degrade the performance because SPs often have very different appearance patterns. In this study, considering that both RL and NAS are very time-consuming, we adopt one-shot and gradient-based NAS in our MARL framework to obtain optimal network architecture for each agent. Specifically, we use the GDAS based method~\cite{dong2019searching} to accelerate the search process while keeping the accuracy. GDAS is a new gradient-based NAS method, which is stable and about 10 times faster than DARTS~\cite{liu2018darts} by only optimizing the sub-graph sampled from the supernet in each iteration.

The agent search process can be defined by search space, search strategy, and performance estimation strategy. Fig.~\ref{fig:framework} shows the designed network architectures, where three agents share 8 cells (5 normal cells and 3 reduce cells) and each agent has its own 4 cells (3 normal cells and 1 reduce cells). Such architecture holds the advantages that agents can not only share knowledge to collaborate with each other, but also learn their specific information as well. The cell search space is then defined as the connection mode and 10 operations including none, $3\times3$ convolution, $5\times5$ convolution, $3\times3$ dilated convolution, $5\times5$ dilated convolution, $3\times3$ separable convolution, $5\times5$ separable convolution, $3\times3$ max pooling, $3\times3$ avg pooling, and skip-connection. For the search strategy, GDAS searches by gradient descent and only updates the sub-graph sampled from the supernet in each iteration. Different batches of data are sampled from the replay buffer according to the prioritized weight and a random way to update network weights $\omega$ and architecture parameters $\alpha$, respectively. We propose a new performance estimation strategy by choosing the optimal architecture parameters $\alpha^*$ when the sum of rewards is maximal on the validation set in all the training epochs and use it to construct the final designed agent $\mathcal{A}_{\alpha^*}$. The main reason of using this strategy is that the loss often oscillates and does not converge in RL.

\subsection{RNN based Agent Collaborative Strategy}
\label{RNN}
Normally, multiple SPs in 3D US have a relatively certain spatial relationship, as shown in Fig.~\ref{fig:intro}. Learning such spatial relationship is critical for automatic multi-plane localization. In~\cite{alansary2018automatic,dou2019agent}, the agents were trained separately for each plane and cannot learn the invaluable spatial relationship among planes. Recently, Vlontzos et al.~\cite{vlontzos2019multiple} proposed a collaborative MARL framework for multi-landmark detection by sharing weights of agent networks across the convolutional layers. However, this strategy is implicit and may not perform well on the multi-plane localization problem. In this study, we propose an RNN based collaborative strategy to learn the spatial layout among planes and improve the accuracy of localization. Specifically, the Bi-direction LSTM (BiLSTM)~\cite{graves2005framewise} is employed in this study, since it can combine the forward and backward information and thus strengthens the communication among agents.

Three agents can collaborate with each other through Q-values. Given the states $s_{t}$ and the network's weights $\omega_t$ in step $t$, the agents output the Q-value sequence set $Q_{t}=(Q_{P_{1}},Q_{P_{2}},Q_{P_{3}})^ \mathrm{T}$, where $Q_{P_{i}}=\{q_{i1},q_{i2},...,q_{i8}\}$ for each plane $P_{i}$. The BiLSTM module then inputs the Q-value sequence $Q_{P_{i}}$ of each plane as its hidden-state of each time-step and outputs the calibrated Q-value sequence set $Q'_{t}=(Q'_{P_{1}},Q'_{P_{2}},Q'_{P_{3}})$ defined by:
\noindent
\begin{equation}
\footnotesize
\begin{split}
\begin{aligned}
Q'_{t} & = \mathcal{H}(Q_{t}
,\stackrel{\rightarrow}{h}_{\tilde{t}-1},\stackrel{\leftarrow}{h}_{\tilde{t}+1};\theta)
\label{equ3}
\end{aligned}
\end{split}
\end{equation}
where $\stackrel{\rightarrow}{h}_{\tilde{t}-1}$ and $\stackrel{\leftarrow}{h}_{\tilde{t}+1}$ are the forward and backward hidden sequences, respectively. $\mathcal{H}$ is the hidden layer function including the input gate, forget gate, output gate and cell, $\theta$ represents the parameters of BiLSTM. 

\section{Experimental Result}
\label{sec:Experimental Results}
\subsubsection{Materials and Implementation Details.}
We validate our proposed method on localizing three uterine SPs in 3D US. Approved by the local Institutional Review Board, 683 volumes were obtained from 476 patients by experts using a US system with an integrated 3D probe. Multiple volumes may be obtained from one patient when her uterus was abnormal or contained IUD. In our dataset, the average volume size is 261$\times$175$\times$277 and the unified voxel size is 0.5$\times$0.5$\times$0.5~$mm^3$. Four experienced radiologists annotated all volumes by manually localizing three SPs and four landmarks (see Fig.~\ref{fig:intro}) in each volume under strict quality control. We randomly split the data into 539 and 144 volumes for training and testing at the patient level to ensure that multiple volumes of one patient belong to the same set.

In this study, we implemented our method in $Pytorch$ and trained the system by Adam optimizer, using a standard PC with a NVIDIA TITAN 2080 GPU. We first obtained optimal agents by setting learning rate as 5e-5 for the weights $\omega$ and 0.05 for the architecture parameters $\alpha$ in 50 epochs (about 2 days). Then we trained the MARL with learning rate=5e-5 and batch size=32 in 100 epochs (about 3 days). The size of replay buffer is set as 15000 and the target network copies the parameters of the current network every 1500 iterations. We further trained the RNN module with hidden size=64 and num layer=2. The starting planes for training the system were randomly initialized around the ground truth within an angle range of $\pm20^{\circ}$ and distance range of $\pm4mm$, according to the average error by landmark-aware alignment~\cite{dou2019agent}. Besides, the step sizes of angle and distance in each 
update iteration are set as $0.5^{\circ}$ and $0.1mm$.

\begin{table}
\scriptsize
\caption{Quantitative evaluation of plane localization.}\label{tab1}
\begin{center}
\begin{tabular}{c|c|c|c|c|c|c|c|c}
\toprule
& Metrics & SARL & MARL & MARL-R & D-MARL & D-MARL-R & G-MARL & G-MARL-R \\
\hline
&Ang($^{\circ}$) & {9.68$\pm$9.63}&  {10.37$\pm$10.18}&{8.48$\pm$9.31}&{8.77$\pm$8.83}&{7.94$\pm$9.15}&{9.60$\pm$8.97}&\textcolor{blue}{7.05$\pm$8.24}\\

S & Dis(mm)&{2.84$\pm$3.49}&\textcolor{blue}{2.00$\pm$2.38}&{2.41$\pm$3.19} &{2.13$\pm$3.01}&{2.03$\pm$3.01}&{2.29$\pm$3.03} & {2.21$\pm$3.21}\\

& SSIM  & {0.88$\pm$0.06}&{0.89$\pm$0.06}&{0.89$\pm$0.07}&{0.88$\pm$0.05}&{0.89$\pm$0.06}&{0.87$\pm$0.12}&\textcolor{blue}{0.90$\pm$0.06}\\
\hline
& Ang($^{\circ}$)& {9.53$\pm$8.27}&{9.30$\pm$8.87}&{8.87$\pm$7.37}&{9.03$\pm$7.91}&{8.57$\pm$7.99}&{8.71$\pm$9.01}&\textcolor{blue}{8.62$\pm$8.07}\\

T & Dis(mm) & {3.17$\pm$2.58}&{2.99$\pm$2.61}&{2.69$\pm$2.49}&\textcolor{blue}{2.01$\pm$2.18}&{2.22$\pm$2.31}&{2.37$\pm$2.36}&{2.36$\pm$2.53}\\
& SSIM & {0.75$\pm$0.10}&{0.72$\pm$0.10}&{0.71$\pm$0.11}&{0.74$\pm$0.12}&{0.75$\pm$0.13}&\textcolor{blue}{0.75$\pm$0.12}&{0.74$\pm$0.13}\\
\hline
& Ang($^{\circ}$) & {8.00$\pm$6.76}&{7.14$\pm$6.67}&{7.17$\pm$6.13}&{7.13$\pm$1.35}&{6.21$\pm$7.11}&{7.21$\pm$6.60}&\textcolor{blue}{5.93$\pm$7.05}\\
C & Dis(mm) & {1.46$\pm$1.40}&{1.53$\pm$1.58}&{1.22$\pm$1.27}&{1.35$\pm$1.42}&{0.96$\pm$1.17}&{1.39$\pm$1.39}&\textcolor{blue}{0.89$\pm$1.15}\\
& SSIM & {0.69$\pm$0.09}&{0.67$\pm$0.09}&{0.68$\pm$0.09}&{0.68$\pm$0.10}&{0.73$\pm$0.12}&{0.68$\pm$0.10}&\textcolor{blue}{0.75$\pm$0.12}\\
\hline
& Ang($^{\circ}$) & {9.07$\pm$8.34}&{8.94$\pm$8.79}&{8.17$\pm$8.13}&{8.31$\pm$8.02}&{7.57$\pm$8.19}&{8.51$\pm$7.98}&\textcolor{blue}{7.20$\pm$7.89}\\
Avg & Dis(mm) & {2.49$\pm$2.74}&{2.17$\pm$2.31}&{2.11$\pm$2.53}&{1.83$\pm$2.32}&\textcolor{blue}{1.74$\pm$2.37}&{2.02$\pm$2.40}&{1.82$\pm$2.54}\\
& SSIM & {0.77$\pm$0.12}&{0.76$\pm$0.13}&{0.76$\pm$0.13}&{0.77$\pm$0.12}&{0.79$\pm$0.13}&{0.77$\pm$0.13}&\textcolor{blue}{0.80$\pm$0.13}\\
\bottomrule
\end{tabular}
\end{center}
\end{table}
\subsubsection{Quantitative and Qualitative Analysis.}
We evaluated the performance in terms of spatial and content similarities for localizing uterine mid-sagittal (S), transverse (T), and coronal (C) planes by the dihedral angle between two planes (Ang), difference of Euclidean distance to origin (Dis) and Structural Similarity 
\begin{table}
\caption{Model information of compared methods.}\label{information}
\begin{center}
\scriptsize
\begin{tabular}{c|c|c|c|c|c|c|c}
\toprule
& V-MARL & MARL & MARL-R & D-MRAL & D-MARL-R & G-MARL & G-MARL-R\\
\hline
FLOPs(G)&{15.41}&{1.82}&{1.82}&{0.69}&{0.69}&{0.68}&{0.68}\\
\hline
Params(M)&{27.58}&{12.22}&{12.35}&{3.62}&{3.76}&{3.61}&{3.75}\\
\bottomrule
\end{tabular}
\end{center}
\end{table}
Index (SSIM). Ablation study was conducted by comparing the methods including Single Agent RL (SARL), MARL, MARL with RNN (MARL-R), DARTS and GDAS based MARL without and with RNN (D-MRAL, D-MARL-R, G-MARL, G-MARL-R). The landmark-aware registration provided warm starts for all the above methods and the Resnet18 served as network backbone for SARL, MARL and MARL-R instead of the VGG to reduce model parameters while keeping comparable performance.

\begin{figure}[htb]
	\centering
	\includegraphics[width=1.0\linewidth]{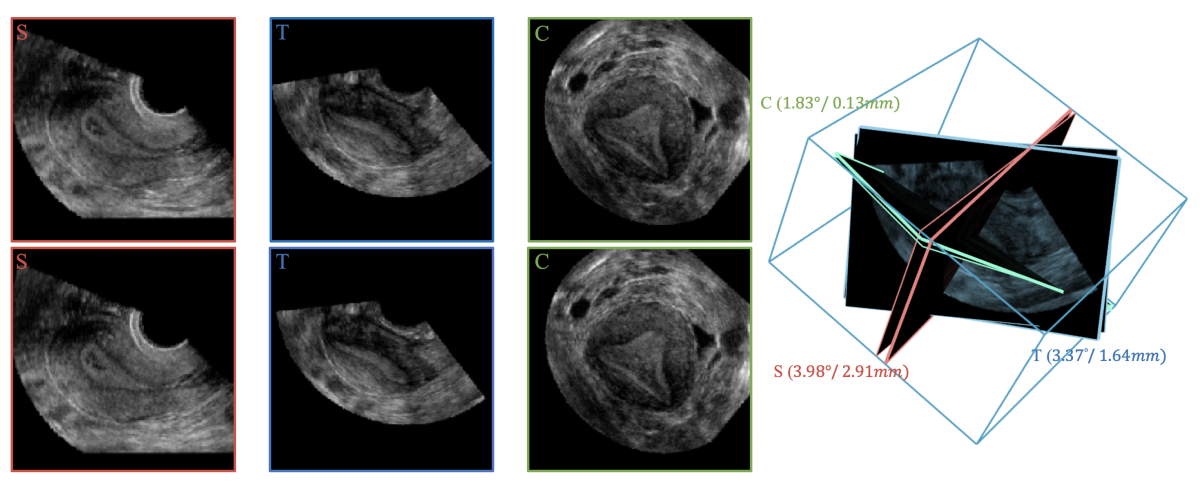}
	\caption{One typical SPs localization result. $Left$: the ground truth (top) and prediction (bottom) of three SPs. $Right$: Spatial differences between prediction and ground truth.}
	\label{fig:result}
\end{figure}

Table~\ref{tab1} lists quantitative results for each plane and average (Avg) values for all planes by different methods. All MARL based methods perform better than the SARL because of the communication among agents. The superior performance of MARL-R, D-MARL and G-MARL compared to MARL shows the efficacy of our proposed agent searching and collaboration methods separately. Among all these methods, our proposed G-MARL-R method achieves the best results, which further illustrates the efficacy of combining these two methods. Table~\ref{information} compares the computational costs (FLOPs) and model sizes (Params) of different methods. The Restnet18-based MARL model has only 12\% FLOPs and 44\% parameters of the VGG-based MARL (V-MARL) model. Meanwhile, NAS-based methods save 63\% FLOPs and 70\% parameters of the MARL. We further compared the GDAS and the DARTS based methods in terms of training time and occupied memory. The former one G-MARL-R (0.04 days/epoch, maximum batch size=32) is much faster and more memory efficient than the D-MARL-R (0.12 days/epoch, maximum batch size=4).

\begin{figure}[htb]
	\centering
	\includegraphics[width=1.0\linewidth]{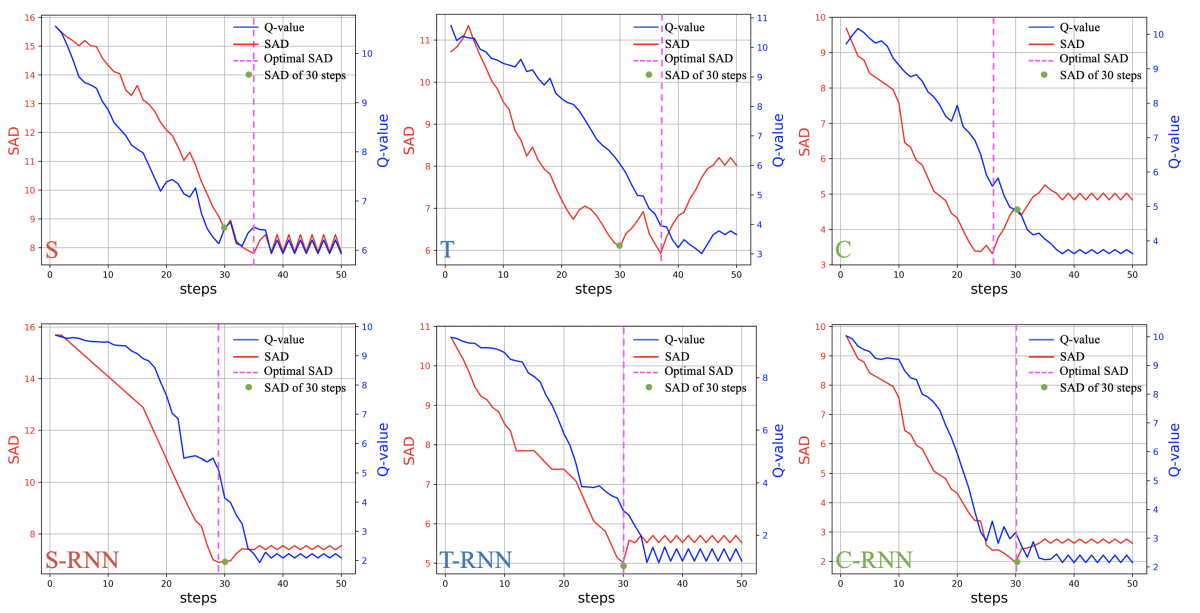}
	\caption{SAD and Q-value curves without (top) and with (bottom) RNN module.}
	\label{fig:SAD}
\end{figure}

Fig.~\ref{fig:result} shows one typical result by the G-MARL-R method. Compared from image content and spatial relationship, our method can accurately localize three SPs in one volume that are very close to the ground truth. Fig.~\ref{fig:SAD} illustrates the Sum of Angle and Distance (SAD) and Q-value curves of the same volume without and with the RNN. Our proposed RNN module greatly reduces the SAD and Q-value, and makes them more stable after a certain number of steps, so it can help accurately localize multiple SPs in 3D US. The optimal SAD is located around step 30, which shows that our termination method based on fixed steps is reasonable and effective.

\section{Conclusion}
\label{sec:conclusion}
We propose a novel MARL framework for multiple SPs localization in 3D US. We use the GDAS-based NAS method to automatically design optimal agents with better performance and fewer parameters. Moreover, we propose an RNN based agent collaborative strategy to learn the spatial relationship among SPs, which is general and effective for establishing strong communications among agents. Experiments on our in-house large dataset validate the efficacy of our method. In the future, we will explore to search Convolutional Neural Network (CNN) and RNN modules together to improve the system performance.
\section{Acknowledgement}
\label{sec:Acknowledgement}
This work was supported by the grant from National Key R$\&$D Program of China (No. 2019YFC0118300), Shenzhen Peacock Plan (No. KQTD2016053112051497, KQJSCX20180328095606003) and Medical Scientific Research Foundation of Guangdong Province, China (No. B2018031).

\bibliographystyle{splncs04}
\bibliography{paper2309}

\begin{thebibliography}{10}
\providecommand{\url}[1]{\texttt{#1}}
\providecommand{\urlprefix}{URL }
\providecommand{\doi}[1]{https://doi.org/#1}

\bibitem{alansary2018automatic}
Alansary, A., Le~Folgoc, L., Vaillant, G., Oktay, O., Li, Y., Bai, W.,
  Passerat-Palmbach, J., Guerrero, R., Kamnitsas, K., Hou, B., et~al.:
  Automatic view planning with multi-scale deep reinforcement learning agents.
  In: International Conference on Medical Image Computing and Computer-Assisted
  Intervention. pp. 277--285. Springer (2018)

\bibitem{chykeyuk2013class}
Chykeyuk, K., Yaqub, M., Noble, J.A.: Class-specific regression random forest
  for accurate extraction of standard planes from 3d echocardiography. In:
  International MICCAI Workshop on Medical Computer Vision. pp. 53--62.
  Springer (2013)

\bibitem{dong2019searching}
Dong, X., Yang, Y.: Searching for a robust neural architecture in four gpu
  hours. In: Proceedings of the IEEE Conference on Computer Vision and Pattern
  Recognition. pp. 1761--1770 (2019)

\bibitem{dou2019agent}
Dou, H., Yang, X., Qian, J., Xue, W., Qin, H., Wang, X., Yu, L., Wang, S.,
  Xiong, Y., Heng, P.A., et~al.: Agent with warm start and active termination
  for plane localization in 3d ultrasound. In: International Conference on
  Medical Image Computing and Computer-Assisted Intervention. pp. 290--298.
  Springer (2019)

\bibitem{graves2005framewise}
Graves, A., Schmidhuber, J.: Framewise phoneme classification with
  bidirectional lstm and other neural network architectures. Neural networks
  \textbf{18}(5-6),  602--610 (2005)

\bibitem{li2018standard}
Li, Y., Khanal, B., et~al.: Standard plane detection in 3d fetal ultrasound
  using an iterative transformation network. In: MICCAI. pp. 392--400. Springer
  (2018)

\bibitem{li2017deep}
Li, Y.: Deep reinforcement learning: An overview. arXiv preprint
  arXiv:1701.07274  (2017)

\bibitem{liu2018darts}
Liu, H., Simonyan, K., Yang, Y.: Darts: Differentiable architecture search.
  arXiv preprint arXiv:1806.09055  (2018)

\bibitem{lorenz2018automated}
Lorenz, C., Brosch, T., Ciofolo-Veit, C., Klinder, T., Lefevre, T., Cavallaro,
  A., Salim, I., Papageorghiou, A.T., Raynaud, C., Roundhill, D., et~al.:
  Automated abdominal plane and circumference estimation in 3d us for fetal
  screening. In: Medical Imaging 2018: Image Processing. vol. 10574, p.
  105740I. International Society for Optics and Photonics (2018)

\bibitem{mnih2015human}
Mnih, V., Kavukcuoglu, K., Silver, D., Rusu, A.A., Veness, J., Bellemare, M.G.,
  Graves, A., Riedmiller, M., Fidjeland, A.K., Ostrovski, G., et~al.:
  Human-level control through deep reinforcement learning. Nature
  \textbf{518}(7540),  529--533 (2015)

\bibitem{ni2014standard}
Ni, D., Yang, X., Chen, X., Chin, C.T., Chen, S., Heng, P.A., Li, S., Qin, J.,
  Wang, T.: Standard plane localization in ultrasound by radial component model
  and selective search. Ultrasound in medicine \& biology  \textbf{40}(11),
  2728--2742 (2014)

\bibitem{ren2020comprehensive}
Ren, P., Xiao, Y., Chang, X., Huang, P.Y., Li, Z., Chen, X., Wang, X.: A
  comprehensive survey of neural architecture search: Challenges and solutions.
  arXiv preprint arXiv:2006.02903  (2020)

\bibitem{ryou2016automated}
Ryou, H., Yaqub, M., Cavallaro, A., Roseman, F., Papageorghiou, A., Noble,
  J.A.: Automated 3d ultrasound biometry planes extraction for first trimester
  fetal assessment. In: International Workshop on Machine Learning in Medical
  Imaging. pp. 196--204. Springer (2016)

\bibitem{schmidt2019offset}
Schmidt-Richberg, A., Schadewaldt, N., Klinder, T., Lenga, M., Trahms, R.,
  Canfield, E., Roundhill, D., Lorenz, C.: Offset regression networks for view
  plane estimation in 3d fetal ultrasound. In: Medical Imaging 2019: Image
  Processing. vol. 10949, p. 109493K. International Society for Optics and
  Photonics (2019)

\bibitem{van2016deep}
Van~Hasselt, H., Guez, A., Silver, D.: Deep reinforcement learning with double
  q-learning. In: Thirtieth AAAI conference on artificial intelligence (2016)

\bibitem{vlontzos2019multiple}
Vlontzos, A., Alansary, A., Kamnitsas, K., Rueckert, D., Kainz, B.: Multiple
  landmark detection using multi-agent reinforcement learning. In:
  International Conference on Medical Image Computing and Computer-Assisted
  Intervention. pp. 262--270. Springer (2019)

\bibitem{wong2015three}
Wong, L., White, N., Ramkrishna, J., J{\'u}nior, E.A., Meagher, S., Costa,
  F.D.S.: Three-dimensional imaging of the uterus: the value of the coronal
  plane. World journal of radiology  \textbf{7}(12), ~484 (2015)

\end{thebibliography}
\end{document}